\documentclass{article}

\usepackage{microtype}
\usepackage{graphicx}
\usepackage{subfigure}
\usepackage{booktabs} 

\usepackage{hyperref}

\usepackage{natbib}

\usepackage{latexsym}
\usepackage{amsfonts}
\usepackage{amsmath}
\usepackage{amsthm}
\usepackage{color}
\usepackage{multirow}
\usepackage{xspace}

\usepackage[accepted]{icml2018}

\icmltitlerunning{Constraining the Dynamics of Deep Probabilistic Models}

\definecolor{mygreen}{rgb}{0.2, 0.7, 0.2}
\definecolor{myorange}{rgb}{0.9, 0.5, 0.0}

\newcommand{\R}{\mathbb{R}}

\newcommand{\LL}{\mathcal{L}}

\newcommand{\norm}{\mathcal{N}}

\newcommand{\fvect}{\mathbf{f}}

\newcommand{\hvect}{\mathbf{h}}

\newcommand{\xvect}{\mathbf{x}}

\newcommand{\yvect}{\mathbf{y}}

\newcommand{\Wvect}{\mathbf{W}}
\newcommand{\tvect}{\mathbf{t}}

\newcommand{\thetavect}{\boldsymbol{\theta}}

\newcommand{\psivect}{\boldsymbol{\psi}}

\newcommand{\Omegavect}{\mathbf{\Omega}}

\newcommand{\name}[1]{{\textsc{#1}}\xspace}

\newcommand{\gp}{\name{gp}}
\newcommand{\gps}{\textsc{gp}s\xspace}
\newcommand{\dgp}{\name{dgp}}
\newcommand{\dgps}{\textsc{dgp}s\xspace}
\newcommand{\dnn}{\name{dnn}}
\newcommand{\dnns}{\textsc{dnn}s\xspace}

\newcommand{\dgpt}{\textsc{dgp-}t\xspace}
\newcommand{\dgpg}{\textsc{dgp-g}\xspace}

\newcommand{\ode}{\name{ode}}
\newcommand{\odes}{\textsc{ode}s\xspace}

\newcommand{\rbf}{\name{rbf}}

\begin{document}

\twocolumn[
\icmltitle{Constraining the Dynamics of Deep Probabilistic Models} 



\icmlsetsymbol{equal}{*}

\begin{icmlauthorlist}
\icmlauthor{Marco Lorenzi}{inria}
\icmlauthor{Maurizio Filippone}{eurecom}
\end{icmlauthorlist}

\icmlaffiliation{inria}{University of Cote d'Azur, INRIA Sophia Antipolis, EPIONE research group, France}
\icmlaffiliation{eurecom}{EURECOM, Department of Data Science, Sophia Antipolis, France}

\icmlcorrespondingauthor{Marco Lorenzi}{marco.lorenzi@inria.fr}
\icmlcorrespondingauthor{Maurizio Filippone}{maurizio.filippone@eurecom.fr}

\icmlkeywords{Deep Gaussian Processes, Probabilistic Deep Learning, Ordinary Differential Equations}

\vskip 0.3in
]



\printAffiliationsAndNotice{\icmlEqualContribution} 

\begin{abstract} 
We introduce a novel generative formulation of deep probabilistic models implementing ``soft'' constraints on their function dynamics.
In particular, we develop a flexible methodological framework where the modeled functions and derivatives of a given order are subject to inequality or equality constraints. 
We then characterize the posterior distribution over model and constraint parameters through stochastic variational inference. 
As a result, the proposed approach allows for accurate and scalable uncertainty quantification on the predictions and on all parameters. 
We demonstrate the application of equality constraints in the challenging problem of parameter inference in ordinary differential equation models, while we showcase the application of inequality constraints on the problem of monotonic regression of count data. 
The proposed approach is extensively tested in several experimental settings, leading to highly competitive results in challenging modeling applications, while offering high expressiveness, flexibility and scalability. 





\end{abstract} 

\section{Introduction}
Modern machine learning methods have demonstrated state-of-art performance in representing complex functions in a variety of applications. 
Nevertheless, the translation of complex learning methods in natural sciences and in the clinical domain is still challenged by the need of interpretable solutions. 
To this end, several approaches have been proposed in order to constrain the solution dynamics to plausible forms such as boundedness \cite{DaVeiga12}, monotonicity \cite{Riihimaki10}, or mechanistic behaviors \cite{alvarez2013linear}. 
This is a crucial requirement to provide a more precise and realistic description of natural phenomena. 
For example, monotonicity of the interpolating function is a common assumption when modeling disease progression in neurodegenerative diseases \cite{lorenzi2017probabilistic,donohue2014estimating}, while bio-physical or mechanistic models are necessary when analyzing and simulating experimental data in bio-engineering \cite{vyshemirsky2007bayesian,konukoglu2011efficient}. 

However, accounting for the complex properties of biological systems in data-driven modeling approaches poses important challenges.
For example, functions are often non-smooth and characterized by nonstationaries which are difficult to encode in ``shallow'' models. 
Complex cases can arise already in classical \ode systems for certain configurations of the parameters, where functions can exhibit sudden temporal changes \cite{goel1971volterra,fitzhugh1955mathematical}. 
Within this context, approaches based on stationary models, even when relaxing the smoothness assumptions, may lead to suboptimal results for both data modeling (interpolation), and estimation of dynamics parameters. 
To provide insightful illustrations of this problem we anticipate the results of Section 4.4.1 and Figure \ref{fig:shallow_vs_deep}. 
Moreover, the application to real data requires to account for the uncertainty of measurements and underlying model parameters, as well as for the -- often large -- dimensionality characterizing the experimental data. 
Within this context, deep probabilistic approaches may represent a promising modeling tool, as they combine the flexibility of deep models with a systematic way to reason about uncertainty in model parameters and predictions. The flexibility of these approaches stems from the fact that deep models implement compositions of functions, which considerably extend the complexity of signals that can be represented with ``shallow'' models \cite{LeCun15}. Meanwhile, their probabilistic formulation introduces a principled approach to quantify uncertainty in parameters estimation and predictions, as well as to model selection problems \cite{Neal96,Ghahramani15}. 

In this work, we aim at extending deep probabilistic models to account for constraints on their dynamics. 
In particular, we focus on a general and flexible formulation capable of imposing a rich set of constraints on functions and derivatives of any order. 
We focus on: i) \emph{equality constraints} on the function and its derivatives, required when the model should satisfy given physical laws implemented through a mechanistic description of a system of interest; and ii) \emph{inequality constraints}, arising in problems where the class of suitable functions is characterized by specific properties, such as monotonicity or convexity/concavity \citep{Riihimaki10}.

In case of equality constraints, we tackle the challenge of parameters inference in Ordinary Differential Equations (\ode).
Exact parameter inference of \ode models is computationally expensive due to the need for repeatedly solving \odes within the Bayesian setting. To this end, previous works attempted to recover tractability by introducing approximate solutions of \odes (see, e.g., \citet{Macdonald15b} for a review).
Following these ideas, we introduce ``soft'' constraints through a probabilistic formulation that penalizes functions violating the \ode on a set of virtual inputs. 
Note that this is in contrast to previous approaches, such as the ones proposed with probabilistic \ode solvers \cite{wheeler2014mechanistic,schober2014probabilistic}, where a given dynamics is strictly enforced to the model posterior.
By deriving a lower bound on the model evidence, we enable the use of stochastic variational inference to achieve end-to-end posterior inference over model and constraint parameters. 
 

In what follows we shall focus on a class of deep probabilistic models implementing a composition of Gaussian processes (\gps) \citep{Rasmussen06} into Deep Gaussian Processes (\dgps) \citep{Damianou13}. 
More generally, our formulation can be straightforwardly extended to probabilistic Deep Neural Networks (\dnns) \citep{Neal96}. 
On the practical side, our formulation allows us to take advantage of  automatic differentiation tools, leading to flexible and easy-to-implement methods for inference in constrained deep probabilistic models. 
As a result, our method scales linearly with the number of observations and constraints. 
Furthermore, in the case of mean-field variational inference, it also scales linearly with the number of parameters in the constraints. 
Finally, it can easily be parallelized/distributed and exploit GPU computing.

Through an in-depth series of experiments, we demonstrate that our proposal achieves state-of-the-art performance in a number of constrained modeling problems while being characterized by attractive scalability properties. 
The paper is organized as follows: Section~2 reports on related work, whereas the core of the methodology is presented in Section~3.  Section~4 contains an in-depth validation of the proposed model against the state-of-the-art. We demonstrate the application of equality constraints in the challenging problem of parameter inference in \ode, while we showcase the application of inequality constraints in the monotonic regression of count data. 
Additional insights and conclusions are given in Section 5.
Results that we could not fit in the manuscript are deferred to the supplementary material.

\section{Related Work}

Equality constraints where functions are enforced to model the solution of \ode systems have been considered in a variety of problems, particularly in the challenging task of accelerated inference of \ode parameters. 
Previous approaches to accelerate \ode parameter optimization involving interpolation date back to \citet{Varah82}. 
This idea has been developed in several ways, including splines, \gps, and Reproducing Kernel Hilbert spaces. 
Works that employ \gps as interpolants have been proposed in \citet{Ramsay07}, \citet{Liang08}, \citet{Calderhead08}, and \citet{Campbell12}.
Such approaches have been extended to introduce a novel formulation to regularize the interpolant based on the \ode system, notably \citet{DondelingerAISTATS13,Barber14}.
An in-depth analysis of the model in \citet{Barber14} is provided by \citet{Macdonald15}.
Recently, \citet{Gorbach17} extended previous works by proposing mean-field variational inference to obtain an approximate posterior over \ode parameters.
Our work improves previous approaches by considering a more general class of interpolants than ``shallow'' \gps, and proposes a scalable framework for inferring the family of interpolating functions jointly with the parameters of the constraint, namely \ode parameters.

Another line of research that builds on gradient matching approaches uses a Reproducing Kernel Hilbert space formulation. 
For example, \citet{Gonzales14} proposes to exploit the linear part of \ode{s} to accelerate the interpolation, while \citet{Niu16} exploits the quadratic dependency of the objective with respect to the parameters of the interpolant to improve the computational efficiency of the \ode regularization. 
Interestingly, inspired by \citet{Calandra16}, the latter approach was extended to handle nonstationarity in the interpolation through warping \citep{Niu17}.
The underlying idea is to estimate a transformation of the input domain to account for nonstationarity of the signal, in order to improve the fitting of stationary \gp interpolants. 
A key limitation of this approach is the lack of a probabilistic formulation, which prevents one from approximating the posterior over \ode parameters.
Moreover, the warping approach is tailored to periodic functions, thus limiting the generalization to more complex signals. 
In our work, we considerably improve on these aspects by effectively modeling the warping through \gps/\dgps that we infer jointly with \ode parameters.   

Inequality constraints on the function derivatives have been considered in several works such as in \citet{meyer2008inference,groeneboom2014nonparametric,mavsic2017shape,Riihimaki10,DaVeiga12,Salzmann10}. 
In particular, the \gp setting provides a solid and elegant theoretical background for tackling this problem; thanks to the linearity of differentiation, both mean and covariance functions of high-order derivatives of \gps can be expressed in closed form, leading to exact formulations for linearly-constrained \gps \citep{DaVeiga12}. 
In case of inequality constraints on the derivatives, instead, this introduces non-conjugacy between the likelihood imposing the derivative constraint and the \gp prior, thus requiring approximations \citep{Riihimaki10}. 
Although this problem can be tackled through sampling schemes or variational inference methods, such as Expectation Propagation \citep{Minka01}, scalability to large dimensions and sample size represents a critical limitation.
In this work, we extend these methods by considering a more general class of functions based on \dgps, and develop scalable inference that makes our method applicable to large data and dimensions.

\section{Methods}

\subsection{Equality constraints in probabilistic modeling}\label{sec:equality}
In this section we provide a derivation of the posterior distribution of our model when we introduce equality constraints in the dynamics.
Let $Y$ be a set of $n$ observed multivariate variables $\yvect_i \in \R^{s}$ associated with measuring times $t$ collected into $\tvect$; the extension where the $n$ variables are measured at different times is notationally heavier but straightforward. 
Let $\fvect(t)$ be a multivariate interpolating function with associated noise parameters $\boldsymbol{\psi}$, and define $F$ similarly to $Y$ to be the realization of $\fvect$ at $\tvect$. 
In this work, $\fvect(t)$ will be either modeled using a \gp, or deep probabilistic models based on \dgps. 
We introduce functional constraints on the dynamics of the components of $\fvect(t)$ by specifying a family of admissible functions whose derivatives of order $h$ evaluated at the inputs $\tvect$ satisfy some given constraint  
$$
\mathcal{C}_{hi} = \left\{\fvect(t) \Biggr\vert \frac{d^h f_i(\tvect)}{dt^h} = \mathcal{H}_{hi}\left(t, \fvect, \frac{d\fvect}{dt}, \ldots, \frac{d^q\fvect}{dt^q}, \thetavect\right) \Biggr\vert_{\tvect}  \right\} \text{.}
$$
Here the constraint is expressed as a function of the input, the function itself, and high-order derivatives up to order $q$. 
The constraint also includes $\thetavect$ as dynamics parameters that should be inferred. 
We are going to consider the intersection of all the constraints for a set of indices $\mathcal{I}$ comprising pairs $(h,i)$ of interest
$$
\mathcal{C} = \bigcap_{(h, i) \in \mathcal{I}} \mathcal{C}_{h i}
$$
To keep the notation uncluttered, and without loss of generality, in the following we will assume that all the terms are evaluated at $\tvect$; we can easily relax this by allowing for the constraints to be evaluated at different sampling points than $\tvect$.
As a concrete example, consider the constraints induced by the Lotka-Volterra \ode system (more details in the experiments section); for this system, $\thetavect = \{\alpha,\beta,\gamma,\delta\}$, and the family of functions is identified by the conditions
$$
\frac{d g_1(t)}{dt}\Biggr\vert_{\tvect} = \mathcal{H}_{11}\left(\fvect(t)\right)\Biggr\vert_{\tvect} = \alpha f_1(\tvect) - \beta f_1(\tvect) f_2(\tvect) \text{,}
$$
$$
\frac{d g_2(t)}{dt}\Biggr\vert_{\tvect} = \mathcal{H}_{12}\left(\fvect(t)\right)\Biggr\vert_{\tvect} = - \gamma f_2(\tvect) + \delta f_1(\tvect) f_2(\tvect) \text{,}
$$
where the products $f_1(\tvect) f_2(\tvect)$ are element-wise.

Denote by $\tilde{F} = \{\fvect_{hi}\}$ the set of realizations of $\fvect$ and of its derivatives at any required order $h$ evaluated at timed $\tvect$.
We define the constrained regression problem through two complementary likelihood-based elements: a data attachment term $p(Y | F, \psivect)$, and a term quantifying the constraint on the dynamics, $p(\mathcal{C} | \tilde{F}, \thetavect, \psivect_D)$, where $\psivect_D$ is the associated noise parameter.
To solve the inference problem, we shall determine a lower bound for the marginal
\begin{align} \label{marginal}
&p(Y, \mathcal{C}  | \tvect, \psivect, \psivect_D) =  \\
&\int p(Y|F, \boldsymbol{\psi}) p(\mathcal{C} | \tilde{F}, \thetavect, \psivect_D)\nonumber  p(F,\tilde{F}| \tvect, \psivect) p(\thetavect) dF d\tilde{F} d\thetavect \text{,}
\end{align}
where 
$$
p(F,\tilde{F} | \tvect, \psivect) =  p(\tilde{F} | F) p(F | \tvect, \psivect).
$$
Note that $\tilde{F}$ is in fact completely identified by $F$.


Equation (\ref{marginal}) requires specifying suitable models for both likelihood and functional constraints. 
This problem thus implies the definition of noise models for both observations and model dynamics.
In the case of continuous observations, the likelihood can be assumed to be Gaussian: 
\begin{equation}
p(Y | F,  \psivect)  = \norm(Y | F,  \Sigma(\psivect)),
\end{equation}
where $\Sigma(\boldsymbol{\psi})$ is a suitable multivariate covariance. 
Extensions to other likelihood functions are possible, and in the experiments we show an application to regression on counts where the likelihood is Poisson with rates equal to the exponential of the elements of $F$.

Concerning the noise model for the derivative observations, we assume independence across the constraints $\mathcal{C}_{hi}$ so that
\begin{equation}
p(\mathcal{C}|\tilde{F}, \theta, \psivect_D) = \prod_{(h,i)\in\mathcal{I}} p(\mathcal{C}_{hi} | \tilde{F}, \theta, \psivect_D).
\end{equation}
We can again assume a Gaussian likelihood:
\begin{equation}
p(\mathcal{C}_{hi}|\tilde{F}, \theta, \psivect_D) = \prod_{t} \norm( \fvect_{hi}(t) | \mathcal{H}_{hi}(t,\tilde{F},\theta), \psivect_D) \text{,}
\end{equation}
or, in order to account for potentially heavy-tailed error terms on the derivative constraints, we can assume a Student-t distribution:
\begin{equation}
p(\mathcal{C}_{hi}|\tilde{F}, \theta, \psivect_D)  = \prod_{t} \mathcal{T}( \fvect_{hi}(t) | \mathcal{H}_{hi}(t,\tilde{F},\theta), \psivect_D, \nu),
\end{equation}
where $\mathcal{T}(z|\mu,\lambda,\nu) \propto \frac{1}{\lambda}[1+\frac{(z-\mu)^2}{\nu\lambda^2}]^{-(\nu+1)/2}$.
We test these two noise models for $\tilde{F}$ in the experiments.

\subsection{Inequality constraints in probabilistic modeling}

In the case of inequality constraints we can proceed analogously as in the previous section. In particular, we are interested in the class of functions satisfying the following conditions:
$$
\mathcal{C}_{hi} = \left\{\fvect(t) \Biggr\vert \frac{d^h f_i(\tvect)}{dt^h} > \mathcal{H}_{hi}\left(t, \fvect, \frac{d\fvect}{dt}, \ldots, \frac{d^q\fvect}{dt^q}, \thetavect\right) \Biggr\vert_{\tvect}  \right\} \text{.}
$$
For example, a monotonic univariate regression problem can be obtained with a constraint of the form $\frac{d f}{dt} > 0 $. 

In this case, the model dynamics can be enforced by a logistic function:
\begin{equation}
p(\mathcal{C}_{hi} | \tilde{F}, \psi_D) = \prod_{j=1}^n \frac{1}{1+\exp(-\psi_D \frac{d f}{dt}(t_j))},
\end{equation}
where the parameter $\psi_D$ controls the strength of the monotonicity constraint.


\subsection{Optimization and inference in constrained regression with \dgps}
After recalling the necessary methodological background, in this section we derive an efficient inference scheme for the model posterior introduced in Section \ref{sec:equality}.

To recover tractability, our scheme leverages on recent advances in modeling and inference in \dgps through approximation via random feature expansions \citep{Rahimi08,Cutajar17}. 
Denoting with $F^{(l)}$ the \gp random variables at layer $l$, an (approximate) \dgp is obtained by composing \gps approximated by Bayesian linear models, $F^{(l)} \approx \Phi^{(l)} W^{(l)}$. 
The so-called random features $\Phi^{(l)}$ are obtained by multiplying the layer input by a random matrix $\Omega^{(l)}$ and by applying a nonlinear transformation $h(\cdot)$.
For example, in case of the standard \rbf covariance, the elements in $\Omega^{(l)}$ are Gaussian distributed with covariance function parameterized through the length-scale of the \rbf covariance. The nonlinearity is obtained through trigonometric functions, $\hvect(\cdot) = (\cos(\cdot), \sin(\cdot))$, while the prior over the elements of $W^{(l)}$  is standard normal. 
As a result, the interpolant becomes a Bayesian Deep Neural Network (\dnn), where for each layer we have weights $\Omega^{(l)}$ and $W^{(l)}$, and activation functions $\hvect(\cdot)$ applied to the input of each layer multiplied by the weights $\Omega^{(l)}$.

\subsubsection{Derivatives in \dgps with random feature expansions }
To account for function derivatives consistently with the theory developed in \citet{Cutajar17}, we need to extend the random feature expansion formulation of \dgps to high-order derivatives. Fortunately, this is possible thanks to the chain rule and to the closure under linear operations of the approximated \gps.  
More precisely, the derivatives of a ``shallow'' \gp model with form $F = \hvect(\tvect \Omega) W$  can still be expressed through linear composition of matrix-valued operators depending on $W$ and $\Omega$ only: $\frac{d{F}}{d t} = \frac{d\hvect(\tvect \Omega)}{dt} W$. 
The computational tractability is thus preserved and the \gp function and derivatives are identified by the same sets of weights $\Omega$ and $W$.
The same principle clearly extends to \dgp architectures where the derivatives at each layer can be combined following the chain rule to obtain the derivatives of the output function with respect to the input.

\subsubsection{Variational lower bound}
In the constrained \dgp setting, we are interested in carrying out inference of the functions $F^{(l)}$ and of the associated covariance parameters at all layers. 
Moreover, we may want to infer any dynamics parameters $\thetavect$ that parameterize the constraint on the derivatives. 
Within this setting, the inference of the latent variables $F^{(l)}$ in the marginal (\ref{marginal}) is generally not tractable.  
Nevertheless, the Bayesian \dnn structure provided by the random feature approximation allows the efficient estimation of its parameters, and the tractability of the inference is thus recovered.  

In particular, let $\Omegavect$, $\Wvect$, and $\psivect$ be the collections of all $\Omega^{(l)}$, $W^{(l)}$, and covariance and likelihood parameters, respectively.
Recalling that we can obtain random features at each layer by sampling the elements in $\Omegavect$ from a given prior distribution, we propose to tackle the inference problem through \emph{variational inference} of the parameters $\Wvect$ and $\boldsymbol{\theta}$.
We could also attempt to infer $\Omegavect$, although in this work we are going to assume them sampled from the prior with fixed randomness, which allows us to optimize covariance parameters using the reparameterization trick (option \name{prior-fixed} in \citet{Cutajar17}).
We also note that we could infer, rather than optimize, $\psivect$; we leave this for future work.


Using Jensen's inequality, the variational approach allows us to obtain a lower bound on the log-marginal likelihood $\LL := \log \left[p(Y, \mathcal{C} | \tvect, \Omegavect, \psivect, \psivect_D)\right]$ of equation (\ref{marginal}), as follows:
\begin{eqnarray}
\mathcal{L} & \geq & E_{q(\Wvect)}\left(\log[p(Y|\Omegavect,\Wvect,\boldsymbol{\psi})] \right)  \nonumber \\
&+& E_{q(\Wvect)q(\boldsymbol{\theta})}\left(\log[p(\mathcal{C}|\Omegavect,\Wvect,\psivect_D, \thetavect)] \right) \nonumber \\
&-& \mathrm{DKL}(q(\Wvect) \| p(\Wvect)) - \mathrm{DKL}(q(\boldsymbol{\theta}) \| p(\boldsymbol{\theta})).
\end{eqnarray}

The distribution $q(\Wvect)$  acts as a variational approximation and is assumed to be Gaussian, factorizing completely across weights and layers $(l)$:
\begin{equation}
q(\Wvect) = \prod_{j,k,l}p(W_{jk}^{(l)}) = \prod_{j,k,l} \norm\left(m_{jk}^{(l)},(s^2)_{jk}^{(l)}\right).
\end{equation}
Extensions to approximations where we relax the factorization assumption are possible.
Similarly, we are going to assume $q(\thetavect)$ to be Gaussian, and will assume no factorization, so that $q(\thetavect) = \norm(\mu_{\theta},\Sigma_{\theta})$.

\section{Experiments}
This section reports an in-depth validation of the proposed method on a variety of benchmarks.
We are going to study the proposed variational framework for constrained dynamics in \dgp models for \ode parameter estimates using equality constraints, and compare it against state-of-the-art methods. 
We will then consider the application of inequality constraints for a regression problem on counts, which was previously considered in the literature of monotonic \gps.

\subsection{Settings for the proposed constrained \dgp}\label{sec:setting}
We report here the configuration that we used across all benchmarks for the proposed method. 
Due to the generally low sample size $n$ used across experiments (in most cases $n < 50$), unless specified otherwise the tests were performed with a two-layer \dgp  $\fvect(t)= \fvect^{(2)} \circ \fvect^{(1)}(t)$, with dimension of the ``hidden'' \gp layer $\fvect^{(1)}(t)$ equal to $2$, and \rbf kernels. 
The length-scale of the \rbf covariances was initialized to $\lambda_0 = \log(t_{\max} - t_{\min})$, while the marginal standard deviation to $\alpha_0 = \log(y_{\max} - y_{\min})$; the initial likelihood noise was set to $\sigma_0^2  = \alpha_0/10^5$. Finally, the initial \ode parameters were set to the value of $0.1$.
The optimization was carried out through stochastic gradient descent with Adaptive moment Estimation (Adam) \cite{kingma2015adam}, through the alternate optimization of i) the approximated posterior over $\Wvect$ and likelihood/covariance parameters ($q(\Wvect)$ and $\psivect$), and ii) likelihood parameters of \ode constraints and the approximate posterior over \ode parameters ($\psivect_D$ and $q(\thetavect)$). 
We note that the optimization of the \ode constraints parameters (the noise and scale parameters for Gaussian and Student-t likelihoods, respectively) is aimed at identifying in a fully data-driven manner the optimal trade-off between data attachment (likelihood term) and regularity (constraints on the dynamics). 
In what follows, \dgpt and \dgpg respectively denote the model tested with Student-t and Gaussian noise models on the \ode constraints.

\subsection{Equality constraints from ODE systems}\label{sec:test_equality}
The proposed framework was tested on a set of \ode systems extensively studied in previous works: Lotka-Volterra \citep{goel1971volterra}, FitzHugh-Nagumo \citep{fitzhugh1955mathematical}, and protein biopathways from \citet{vyshemirsky2007bayesian}. For each experiment, we used the experimental setting proposed in previous studies \citep{Niu16,Macdonald15b}. In particular, for each test, we identified two experimental configurations with increasing modeling difficulty (e.g. less samples, lower signal-to-noise ratio, $\ldots$). A detailed description of the models and testing parameters is provided in the supplementary material.
The experimental results are reported for parameter inference and model estimation performed on $5$ different realizations of the noise.

\subsubsection{Benchmark}
We tested the proposed method against several reference approaches from the state-of-art to infer parameters of \ode systems. 

{\bf{RKG3:}} We tested the method presented in \citet{Niu16} using the implementation in the \texttt{R} package \texttt{KGode}.
This method implements gradient matching, where the interpolant is modeled using functions in Reproducing Kernel Hilbert spaces. 
This approach, for which \ode parameters are estimated and not inferred, was shown to achieve state-of-the-art performance on a variety of \ode estimation problems. 
We used values ranging from $10^{-4}$ to $1$ for the parameter $\lambda$ that the method optimizes using cross-validation.

{\bf{Warp:}} In the \texttt{R} package \texttt{KGode} there is also an implementation of the warping approach presented in \citet{Niu17}.
This method extends gradient matching techniques by attempting to construct a warping of the input where smooth Reproducing Kernel Hilbert spaces-based interpolants can effectively model nonstationary observations. 
The warping attempts to transform the original signal via assumptions on periodicity and regularity conditions.
We used the default parameters and initialized the optimization of the warping function from a period equal to the interval where observations are available.
Similarly to RKG3, \ode parameters are estimated and not inferred.

{\bf{AGM:}} We report results on the Approximate Gradient Matching (AGM) approach in \citet{DondelingerAISTATS13}, implemented in the recently released \texttt{R} package \texttt{deGradInfer}.
AGM implements a population Markov chain Monte Carlo approach tempering from the prior to the approximate posterior of \ode parameters based on an interpolation with \gps.
In the experiments we use $10$ parallel chains and we run them for $10^4$ iterations. 
In the implementation of AGM, the variance of the noise on the observations is assumed known and it is fixed; we expect this to give a slight advantage to this method.

{\bf{MCMC:}} In the \texttt{R} package \texttt{deGradInfer} there is also an implementation of a population Markov chain Monte Carlo sampler where the \ode is solved explicitly. 
In this case too we use $10$ parallel chains that we run for $10^4$ iterations. 
In contrast to AGM, in this implementation, the variance of the noise on the observations is learned together with \ode parameters.

\subsubsection{Results}

\begin{figure}[t]
\vskip 0.2in
\centering
\begin{tabular}{@{}c@{}}
{\bf {\scriptsize Lotka-Volterra }}\\
\vspace{-2mm}
\includegraphics[width=\columnwidth]{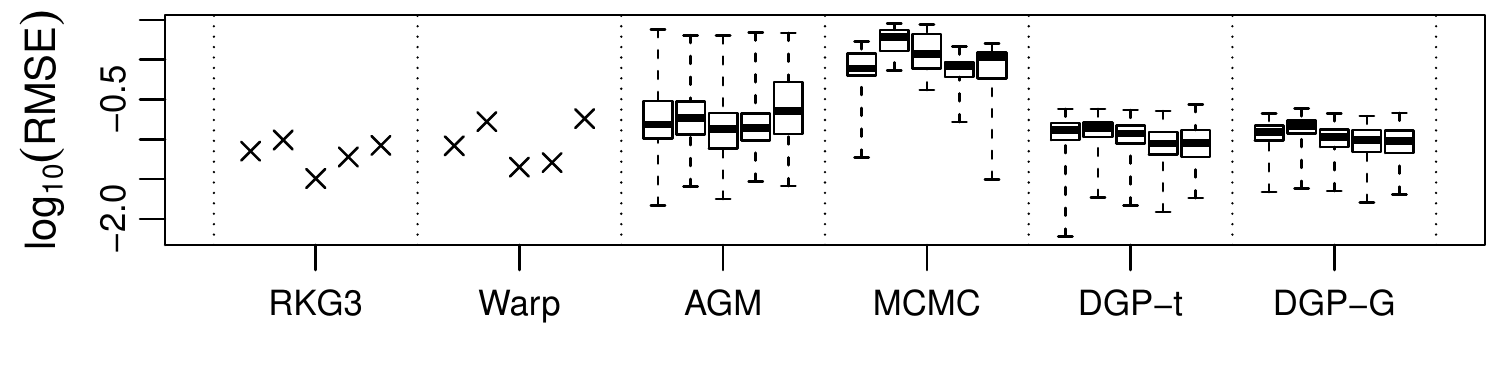} \\
\includegraphics[width=\columnwidth]{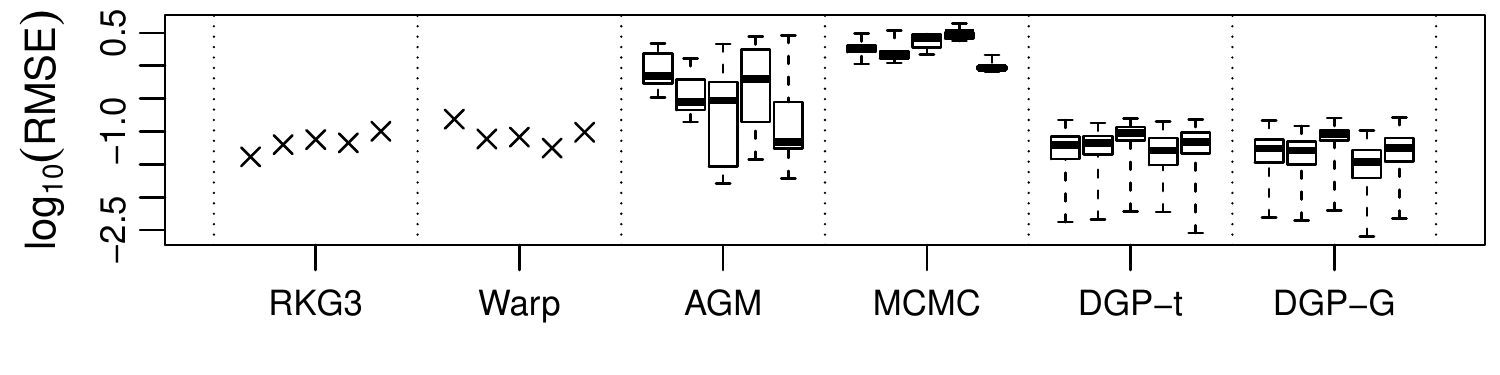}\\
{\bf {\scriptsize FitzHugh-Nagumo }}\\
\vspace{-2mm}
\includegraphics[width=\columnwidth]{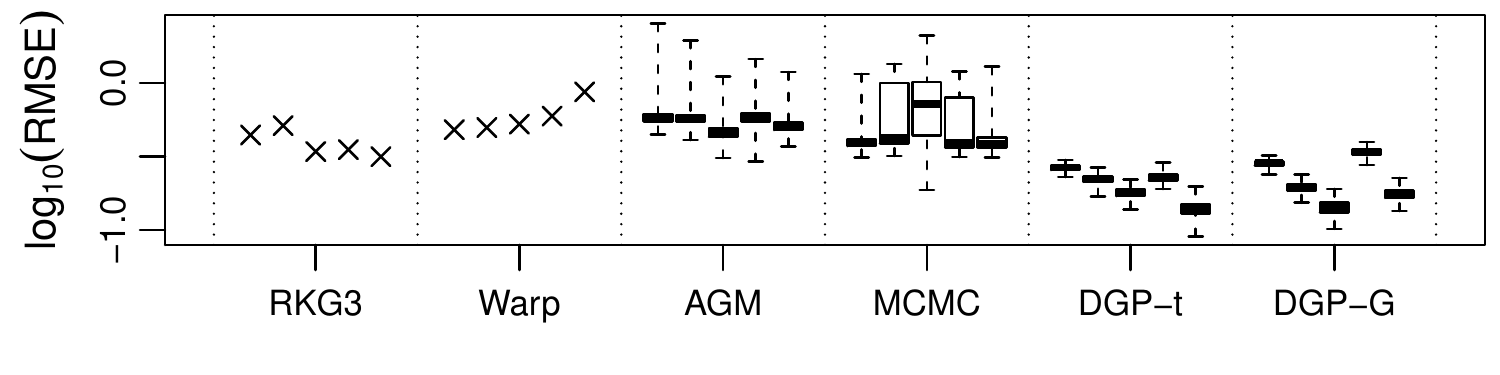} \\
\includegraphics[width=\columnwidth]{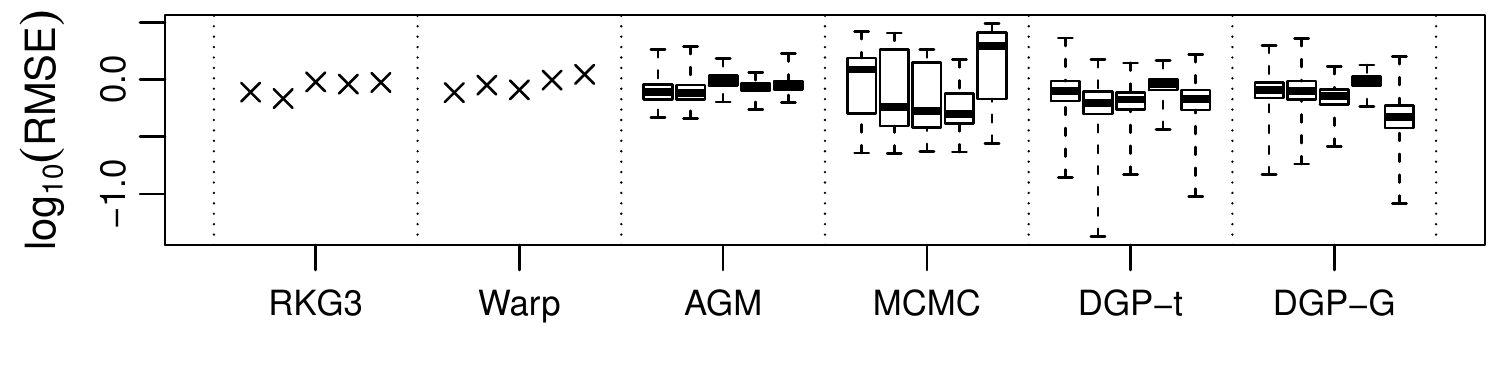} \\
{\bf {\scriptsize Biopathways }}\\
\vspace{-2mm}
\includegraphics[width=\columnwidth]{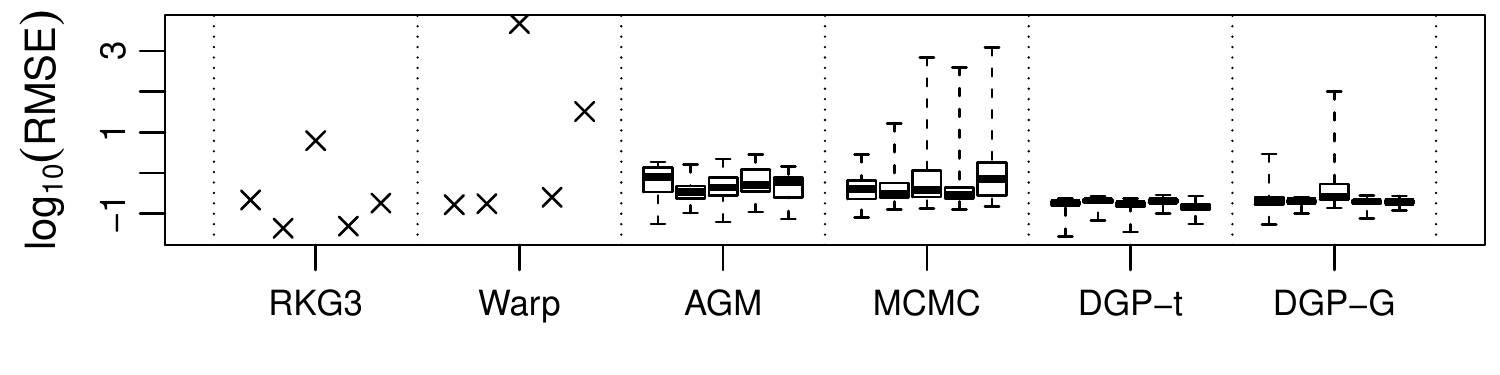}\\
\includegraphics[width=\columnwidth]{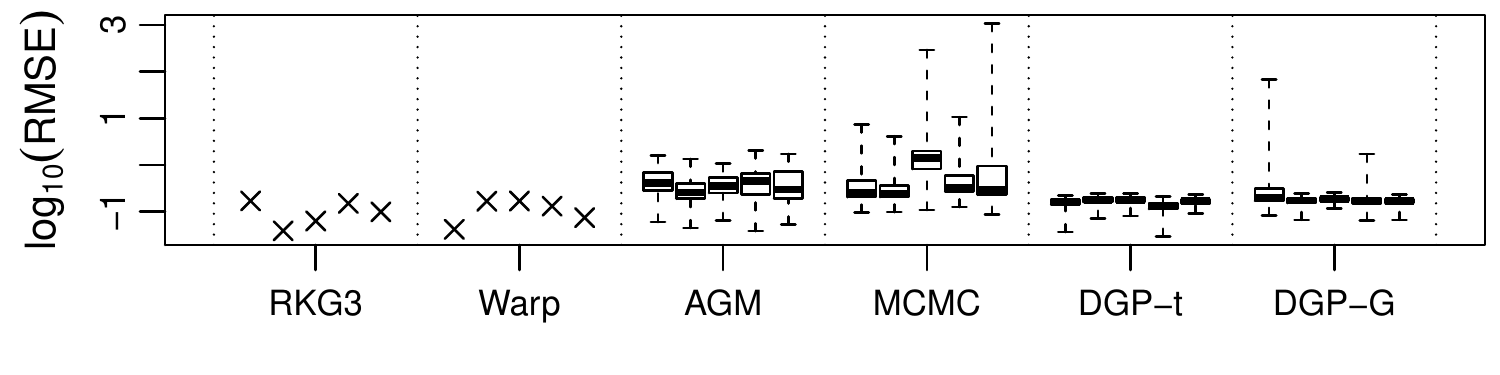}
\label{fig:ode1}
\end{tabular}
\caption{Boxplot of the RMSE on \ode parameters for the three \ode systems considered and for the two experimental settings. 
We report $5$ bars for each method in the plots, corresponding to five different instantiations of the noise.}
\label{fig:ode}
\vskip -0.2in
\end{figure}

Figure \ref{fig:ode1} shows the distribution of the root mean squared error (RMSE) across folds for each experimental setting (see supplement for details). 
We note that the proposed method consistently leads to better RMSE values compared to the reference approaches (except some folds in one of the Fitz-Hugh-Nagumo experiments, according to a Mann-Whitney nonparametric test), and that \dgpt provides more consistent parameter estimates than \dgpg. 
This latter result may indicate a lower sensitivity to outliers derivatives involved in the functional constraint term. 
This is a crucial aspect due to the generally noisy derivative terms of nonparametric regression models. 
The distribution of the parameters for all the datasets tested in this study, which we report in the supplementary material, reveals that, unlike the nonprobabilistic methods RKG3 and WARP, our approach is capable of inferring \ode parameters yielding meaningful uncertainty estimation. 

\subsection{Scalability test - large $n$}\label{sec:large_n}

\begin{figure}[t]
\vskip 0.2in
\begin{center}
{\bf {\scriptsize Lotka-Volterra}} \\
\includegraphics[width=7cm]{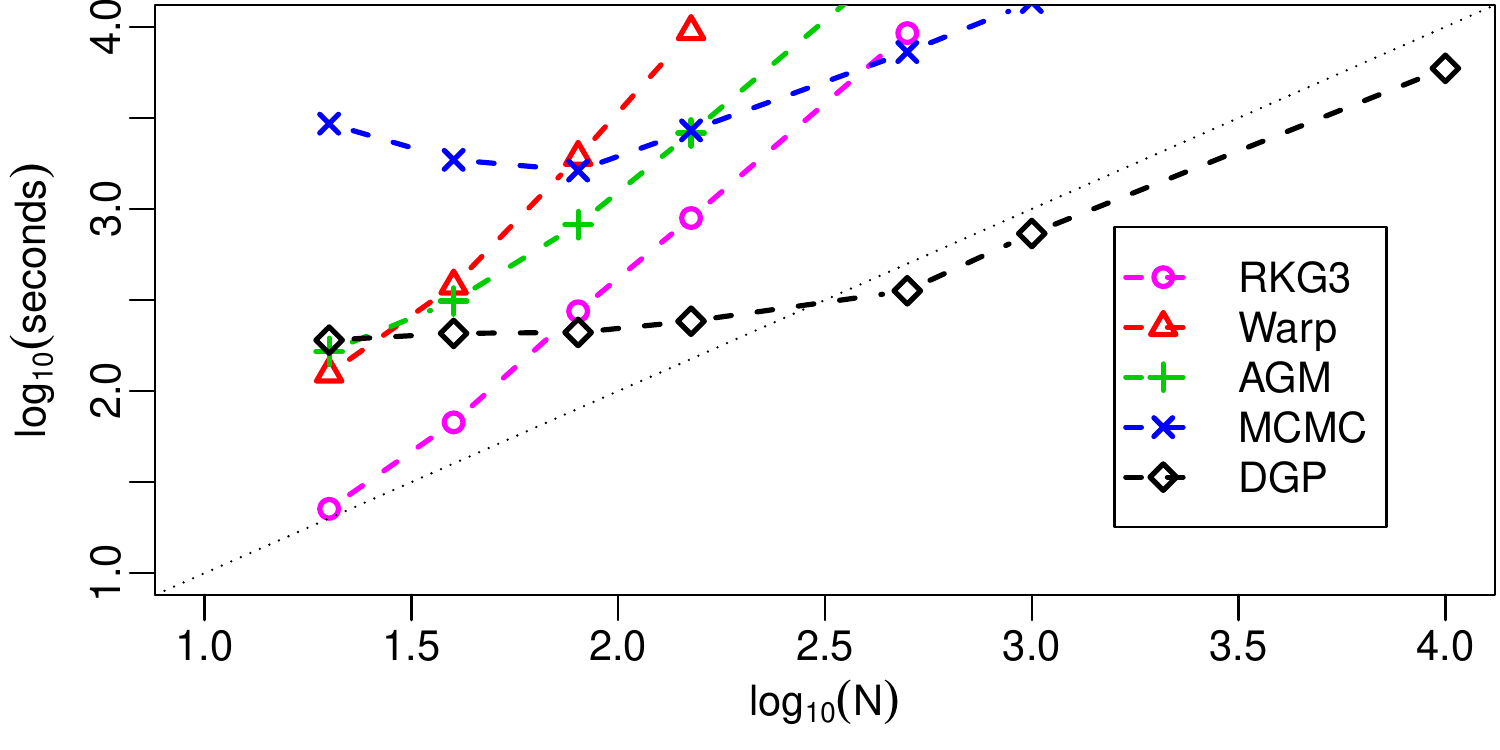}
\caption{Execution time vs sample size -- Lotka-Volterra \ode.}
\label{fig:large_n}
\end{center}
\vskip -0.2in
\end{figure}

We tested the scalability of the proposed method with respect to sample size.
To this end, we repeated the test on the Lotka-Volterra system with $n= 20, 40, 80,150,500, 10^3$, and $10^4$ observations. 
For each instance of the model, the execution time was recorded and compared with the competing methods. All the experiments were performed on a $1.3 \mathrm{GHz}$ Intel Core i5 MacBook.
The proposed method scales linearly with $n$ (Figure \ref{fig:large_n}), while it has an almost constant execution when $n<500$; we attribute this effect to overheads in the framework we used to code our method.
For small $n$, the running time of our method is comparable with competing methods, and it is considerably faster in case of large $n$.

\subsection{Scalability test - large $s$}\label{sec:large_s}

In order to assess the ability of the framework to scale to a large number of {\ode}s, we tested our method on the Lorenz96 system with increasing number of equations, $s=125$ to $s=1000$ \citep{lorenz1998optimal}. To the best of our knowledge, the solution of this challenging problem via gradient matching approaches has only been previously attempted in \citet{Gorbach17}. 
We could not find an implementation of their method to carry out a direct comparison, so we are going to refer to the results reported in their paper. 
The system consists of a set of drift states functions $\left(f_1(\xvect(t),\theta), f_2(\xvect(t),\theta),\ldots, f_s(\xvect(t),\theta)\right)$ recursively linked by the relationship:
$$
f_i(\xvect(t),\theta) = (x_{i+1}(t) - x_{i-2}(t)) x_{i-1}(t) - x_{i}(t) + \theta \text{,}
$$
where $\theta \in \R$ is the drift parameter.
Consistently with the setting proposed in \citet{Gorbach17,vrettas2015variational}, we set $\theta = 8$ and generated 32 equally spaced observations over the interval $[0,4]$ seconds, with additive Gaussian noise $\sigma^2 = 1$. We performed two tests by training (i) on all the states, and (ii) by keeping one third of the states as unobserved, and by applying our method to identify model dynamics on both observed and unobserved states.

\begin{figure}[t]
\vskip 0.2in
\begin{center}
\includegraphics[width=7cm]{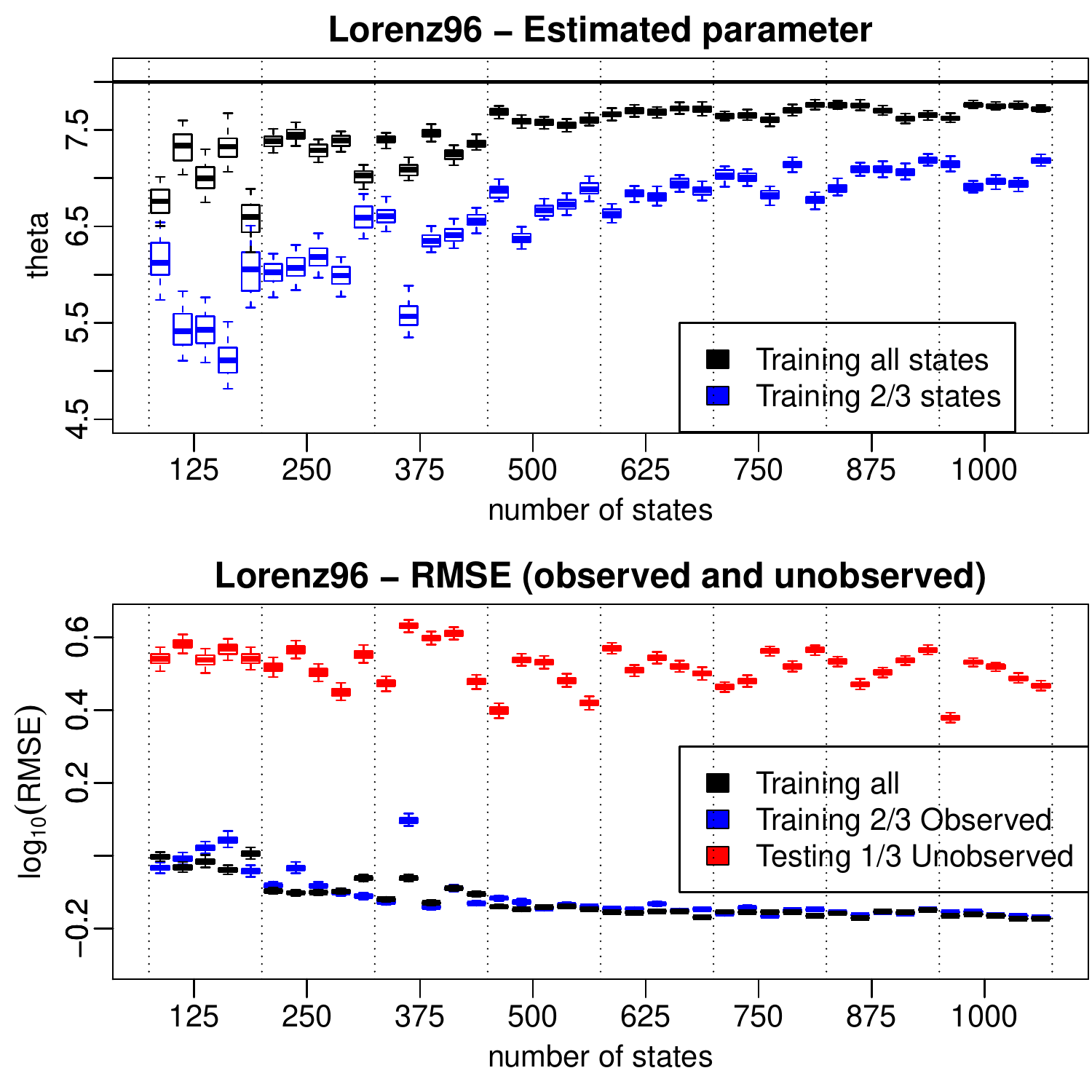}
\caption{Top: parameter estimates in the different folds when training on all (black) or only on 2/3 (blue) of the states. The ground truth is indicated by the top horizontal bar ($\theta = 8$). Bottom: RMSE on the \ode curves fitting when training on all the states (black), and on the observed (blue) and unobserved (red) states when training on 2/3 of the states only. }
\label{fig:Lorenz96}
\end{center}
\vskip -0.2in
\end{figure}

\begin{figure}[t]
\vskip 0.2in
\begin{center}
{\bf {\scriptsize Lorenz96 - Observed }}
\includegraphics[width=\columnwidth]{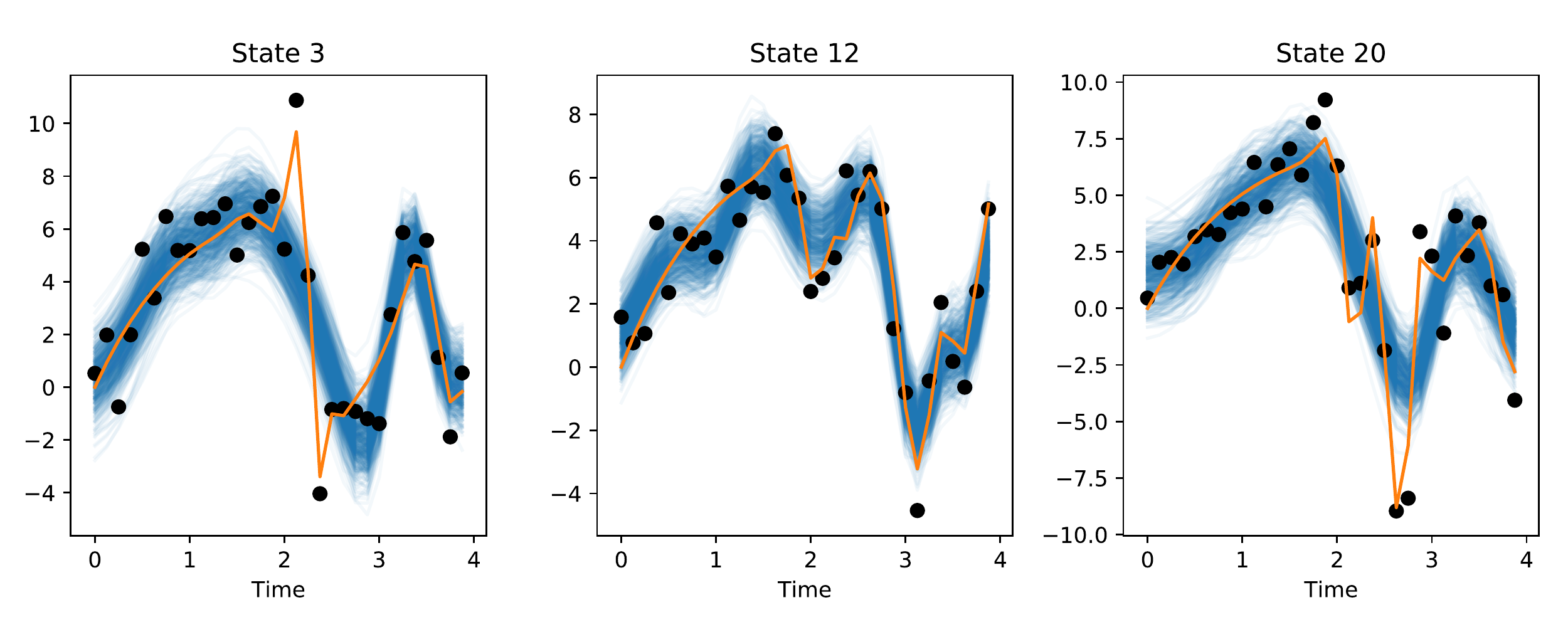}
{\hspace{0.3cm}\bf {\scriptsize Lorenz 96 - Unobserved \\}}
\includegraphics[width=\columnwidth]{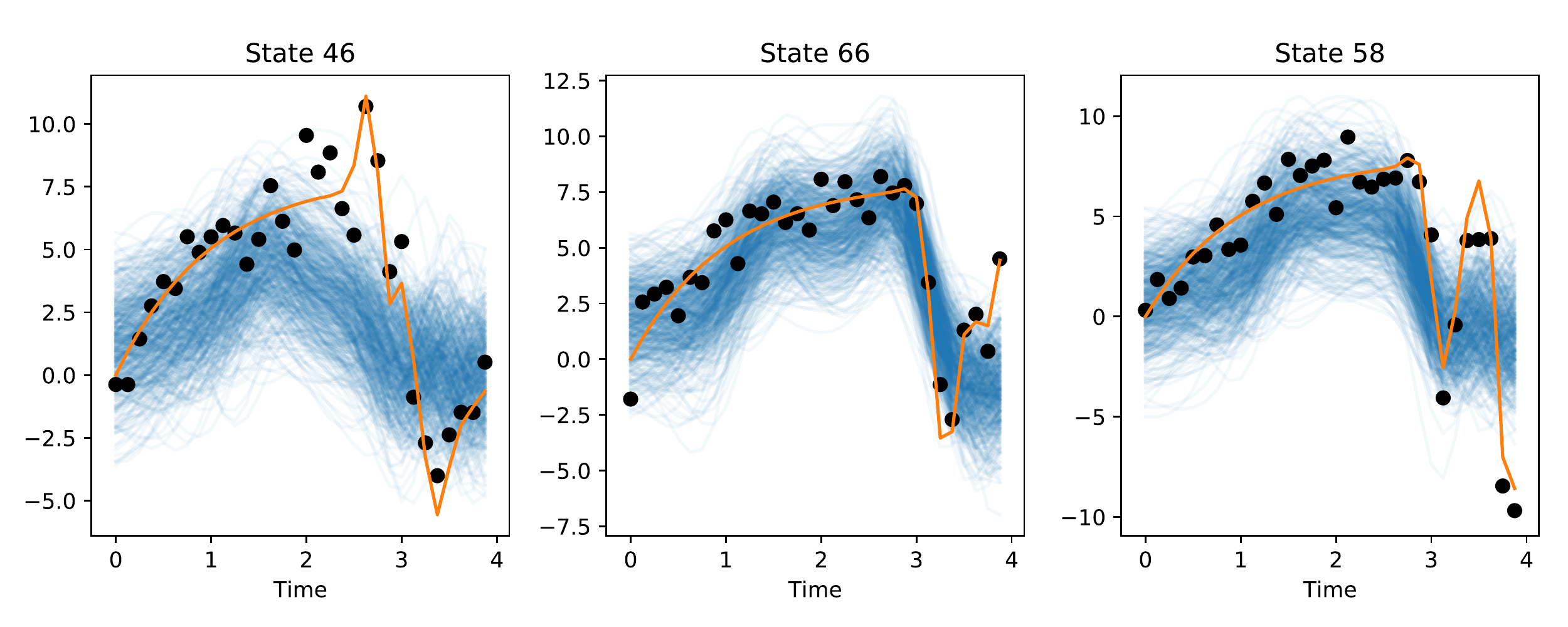}
\caption{Model fit in Lorenz96. Randomly sampled observed (top) vs unobserved (bottom) states for $s=125$ \odes. Orange lines and black dots represent respectively the ground truth dynamics and noisy sample points. The blue lines are realizations of the \dgp. }
\label{fig:Lorenz96_plots}
\end{center}
\vskip -0.2in
\end{figure}

Figure \ref{fig:Lorenz96} shows the average RMSE in the different experimental settings. As expected, the modeling accuracy is sensibly higher when trained on the full set of equations. Moreover, the RMSE is lower on observed states compared to unobserved ones. This is confirmed by visual inspection of the modeling results for sample training and testing states (Figure \ref{fig:Lorenz96_plots}). 
The observed states are generally associated with lower uncertainty in the predictions and by an accurate fitting of the solutions (Figure \ref{fig:Lorenz96_plots}, top). 
The model still provides remarkable modeling results on unobserved states (Figure \ref{fig:Lorenz96_plots}, bottom), although with decreased accuracy and higher uncertainty.
We are investigating the reasons for the posterior distribution over $\theta$ not covering the true value of the parameter across different experimental conditions. 

\subsubsection{Deep vs shallow}\label{sec:deep_vs_shallow}


We explore here the capability of a \dgp to accommodate for the data nonstationarity typical of \ode systems. 
In particular, the tests are performed in two different settings with large and small sample size $n$. 
By using the same experimental setting of Section \ref{sec:setting}, we sampled $80$ and $1000$ points, respectively, from the FitzHugh-Nagumo equations. 
The data is modeled with \dgps composed by one (``shallow'' \gp), two and three layers, all with \rbf covariances.

\begin{table}[ht]
	\caption{Shallow and deep \gp models under different experimental conditions in FitzHugh-Nagumo equations. Best results are highlighted in bold.}
	\label{tab:shallow_vs_deep}
        \vskip 0.15in
	\begin{center}
\begin{small}
\begin{sc}
	\begin{tabular}{c|c|c|c|c|c|c|c} 
\toprule
        \multicolumn{8}{c}{average rmse across parameters}\\
\midrule
        & \multicolumn{3}{c|}{\rbf} & \multicolumn{4}{c}{Mat\'ern}\\
        & \multicolumn{3}{c|}{Layers} & \multicolumn{4}{c}{$\nu$} \\		
       	$n$ & $1$ & $2$ & $3$ & $1/2$ & $1$ & $3/2$ & $5/2$ \\		
\midrule
		80   & 0.86 &  0.85 & 2.16 & 0.97 & {\bf 0.72} & 0.79 & 0.77 \\
		1000 & 0.66 &  {\bf 0.52} & 0.53 & 0.87 & 0.63 & 0.70 & 0.74 \\
\midrule
		\multicolumn{8}{c}{ data fit rmse } \\
\midrule
		80   & 0.23 &  {\bf 0.19} & 0.42 & 0.23 & 0.20 & 0.23 & 0.25 \\
		1000 & 0.22 &  {\bf 0.17} & 0.19 & 0.21 & 0.21 & 0.19 & 0.24 \\
\bottomrule
	\end{tabular}
\end{sc}
\end{small}
        \end{center}
        \vskip -0.1in
\end{table}

\begin{figure}[t]
\centering
\includegraphics[width=\columnwidth]{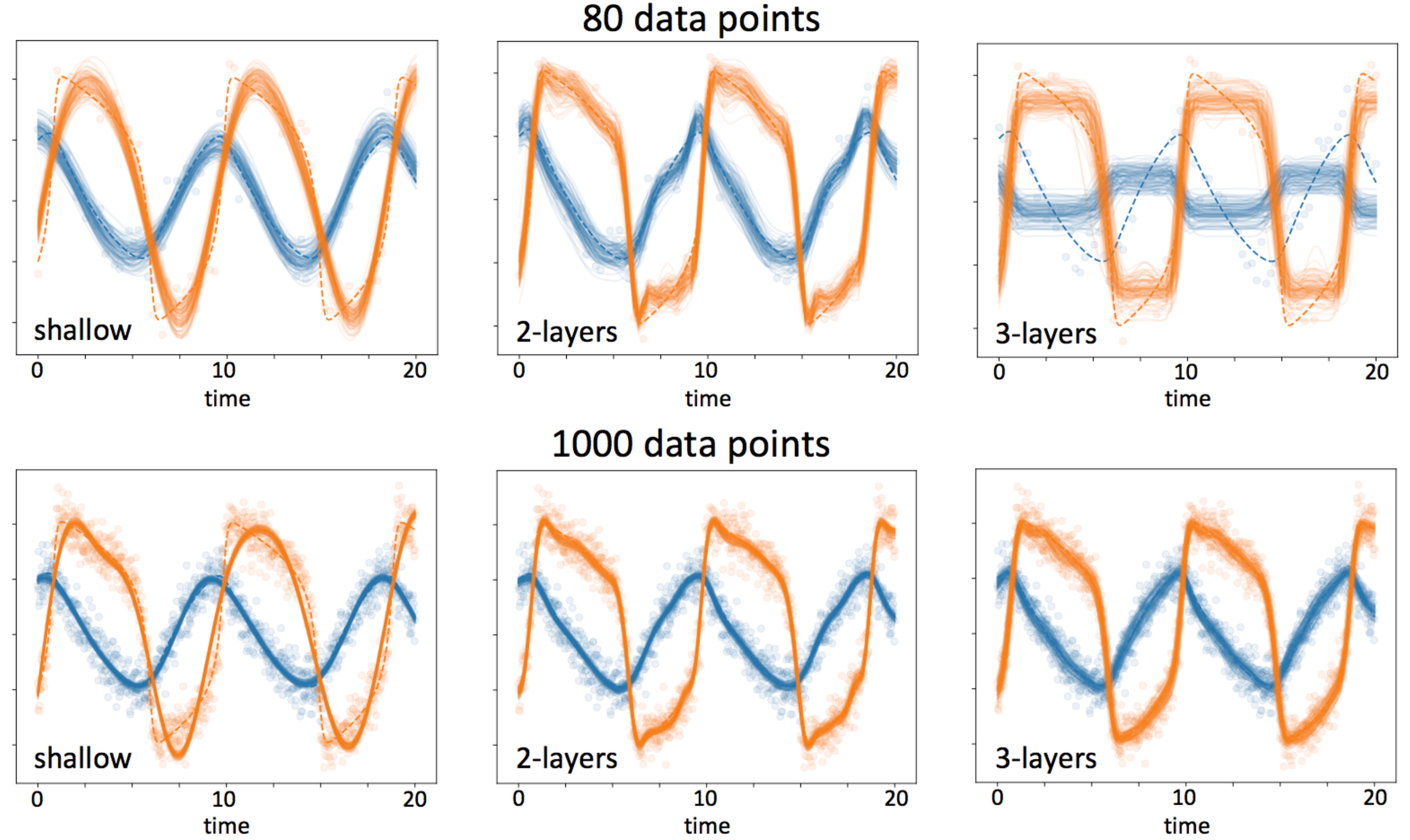}
\caption{{Modeling FitzHugh-Nagumo equations with \gp and \dgp. A deep model provides a more accurate description of data nonstationarity and associated dynamics (Table \ref{tab:shallow_vs_deep}). Training points are denoted with circles; the ground truth trajectory is represented by the dashed line. Top: $N = 80$; Bottom: $N=1000$. From left to right: Shallow \gp, two-layers and three-layers  \dgp.}}\label{fig:shallow_vs_deep} 
\vskip -0.2in
\end{figure}

Figure \ref{fig:shallow_vs_deep} shows the modeling results obtained on the two configurations.  
We note that the shallow \gp consistently underfits the complex dynamics producing smooth interpolants. 
On the contrary, {\dgp}s provide a better representation of the nonstationarity. As expected, the three-layer \dgp leads to sub-optimal results in the low-sample size setting. 
Furthermore, in order to motivate the importance of nonstationarity, which we implement through \dgps, we further compared against shallow \gps with lower degrees of smoothness through the use of Mat\'ern covariances with degrees $\nu = 1/2, 1, 3/2, 5/2$. 

The overall performance in parameter estimation and data fit is reported in Table \ref{tab:shallow_vs_deep}. 
According to the results, a two-layer \dgp provides the best solution overall in terms of modeling accuracy and complexity.
Interestingly, the Mat\'ern covariance, with an appropriate degree of smoothness, achieves superior performance in parameter estimation in case of low sample size. 
However, the nonstationarity implemented by the \dgp outperforms the stationary Mat\'ern in the data fit, as well as in the parameter estimation when the sample size is large. For an illustration of the data fit of the Mat\'ern \gp we refer the reader to the supplementary material. 
Crucially, these results indicate that our approach provides a practical and scalable way to learn nonstationarity within the framework of variational inference for deep probabilistic models.

\subsection{Inequality constraints}\label{sec:test_inequality}

\begin{figure}[t]
\vskip 0.2in
\begin{center}
\includegraphics[width=\columnwidth]{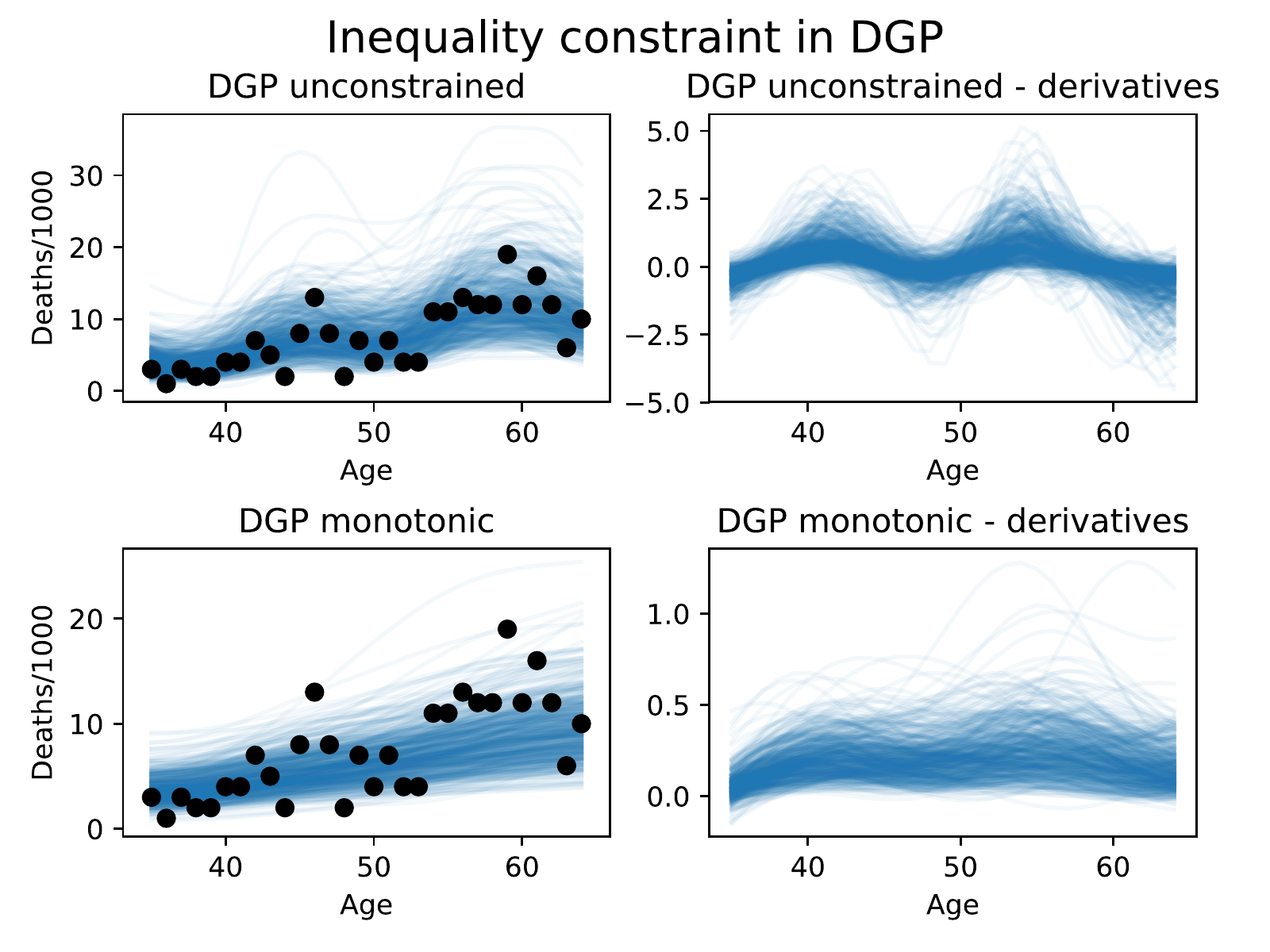}
\caption{\gp with Poisson likelihood: unconstrained (top) and monotonic (bottom). Black dots: observations from \cite{broffitt1988increasing}. Blue lines: \gp realizations. }
\label{fig:monotonic}
\end{center}
\vskip -0.2in
\end{figure}

We conclude our experimental validation by applying monotonic regression on counts as an illustration of the proposed framework for inequality constrains in \dgp models dynamics.
We applied our approach to the mortality dataset from \cite{broffitt1988increasing}, with a two-layer \dgp initialized with an analogous setting to the one proposed in Section \ref{sec:setting}. 
In particular, the sample rates were modeled with with a Poisson likelihood of the form $p(y_i|\mu_i) = \frac{\exp(-\mu_i) \mu_i^{y_i}}{y_i!}$, and link function $\mu_i = \exp(f(t_i))$. 
Monotonicity on the solution was strictly enforced by setting $\psi_D = 5$. 
Figure \ref{fig:monotonic} shows the regression results without (top) and with (bottom) monotonicity constraint. The effect of the constraint on the dynamics can be appreciated by looking at the distribution of the derivatives (right panel). In the monotonic case the \gp derivatives lie on the positive part of the plane. This experiment leads to results compatible with those obtained with the monotonic \gp proposed in \cite{Riihimaki10}, and implemented in the GPstuff toolbox \cite{vanhatalo2013gpstuff}.
However, our approach is characterized by appealing scalability properties and can implement monotonic constraints on \dgps, which offer a more general class of functions than \gps. 

\section{Conclusions}
We introduced a novel generative formulation of deep probabilistic models implementing ``soft'' constraints on functions dynamics.
The proposed approach was extensively tested in several experimental settings, leading to highly competitive results in challenging modeling applications, and favorably comparing with the state-of-the-art in terms of modeling accuracy and scalability. 
Furthermore, the proposed variational formulation allows for a meaningful uncertainty quantification of both model parameters and predictions. 
This is an important aspect intimately related to the application of our proposal in real scenarios, such as in biology and epidemiology, where data is often noisy and scarce.

Although in this study we essentially focused on the problem of \ode parameters inference and monotonic regression, the generality of our approach enables several other applications that will be subject of future investigations. 
We will focus on the extension to manifold valued data, such as spatio-temporal observations represented by graphs, meshes, and 3D volumes, occurring for example in medical imaging and system biology. 

\section*{Acknowledgements} 
This work has been supported by the French government, through the UCAJEDI Investments in the Future project managed by the National Research Agency (ANR) with the reference number ANR-15-IDEX-01 (project Meta-ImaGen).
MF gratefully acknowledges support from the AXA Research Fund.

This work is dedicated to Mattia Filippone.



\appendix












\onecolumn 

\section{Description of the \ode systems considered in this work}

{\bf{Lotka-Volterra}} \citep{goel1971volterra}. This \ode describes a two-dimensional process with the following dynamics: 
\begin{eqnarray}	
\frac{d f_1}{dt} = \alpha f_1 - \beta f_1 f_2; \qquad \frac{d f_2}{dt} =  - \gamma f_2 + \delta f_1 f_2, \nonumber
\end{eqnarray}
and is identified by the parameters $\boldsymbol{\theta} = \{\alpha,\beta,\gamma,\delta\}$. Following \cite{Niu16} we generated a ground truth from numerical integration of the system with parameters $\thetavect = \{0.2,0.35,0.7,0.4\}$ over the interval $[0,30]$ and with initial condition $[1,2]$. We generated two different configurations, composed by respectively 34 and 51 observations sampled at uniformly spaced points, and corrupted by zero mean Gaussian noise with standard deviation $\sigma = 0.25$ and $\sigma =0.4$ respectively.

{\bf{FitzHugh-Nagumo}} \citep{fitzhugh1955mathematical}. This system describes a two-dimensional process governed by 3 parameters, $\thetavect = \{a, b, c\}$:
\begin{eqnarray}	
\frac{d f_1}{dt} = c (f_1 - b \frac{(f_1)^3}{3} +  f_2); \qquad \frac{d f_2}{dt} =  - \frac{1}{c}( f_1 - a + b * f_2). \nonumber
\end{eqnarray}
We reproduced the experimental setting proposed in \cite{Macdonald15b}, by generating a ground truth with $\thetavect = \{3,0.2,0.2\}$, and by integrating the system numerically with initial condition $[-1,1]$. We created two scenarios; in the first one, we sampled 401 observations at equally spaced points within the interval $[0,20]$, while in the second one we sampled only 20 points. In both cases we corrupted the observations with zero-mean Gaussian noise with $\sigma = 0.5$. 

{\bf{Biopathways}} \citep{vyshemirsky2007bayesian}. These equations describe a five-dimensional process associated with 6 parameters $\thetavect = \{k_1, k_2, k_3, k_4, V, K_m\}$ as follows:
\begin{eqnarray}	
\frac{d f_1}{dt} &=& -k_1f_1 - k_2 f_1f_3 + k_3 f_4; \nonumber\\
\frac{d f_2}{dt} &=&  k_1f_1; \nonumber\\
\frac{d f_3}{dt} &=&  -k_2 f_1 f_3 + k_3 f_4 + \frac{V f_5}{K_m+f_5}; \nonumber\\
\frac{d f_4}{dt} &=&  k_2f_1f_3 - k_3 f_4 - k_4 f_4 ; \nonumber\\
\frac{d f_5}{dt} &=&  k_4 f_4 - \frac{V f_5}{K_m+f_5} . \nonumber
\end{eqnarray}
We generated data  by sampling 15 observations at times $\tvect = \{0,1,2,4,5,7,10,15,20,30,40,50,60,80,100\}$ \cite{Macdonald15b}. The \ode parameters were set to $\thetavect = \{k_1=0.07,k_2=0.6,k_3=0.05,k_4=0.3,V=0.017,K_m=0.3\}$, and the initial values were $[1,0,1,0,0]$. We generated two different scenarios, by adding Gaussian noise with $\sigma^2 = 0.1$ and $\sigma^2 = 0.05$, respectively.

\newpage 

\section{Detailed results of the benchmark on \ode parameter inference}

\begin{figure*}[h!]
\vskip 0.2in
\begin{center}
{\bf {\scriptsize Lotka-Volterra }}
\includegraphics[width=\columnwidth, height=2cm]{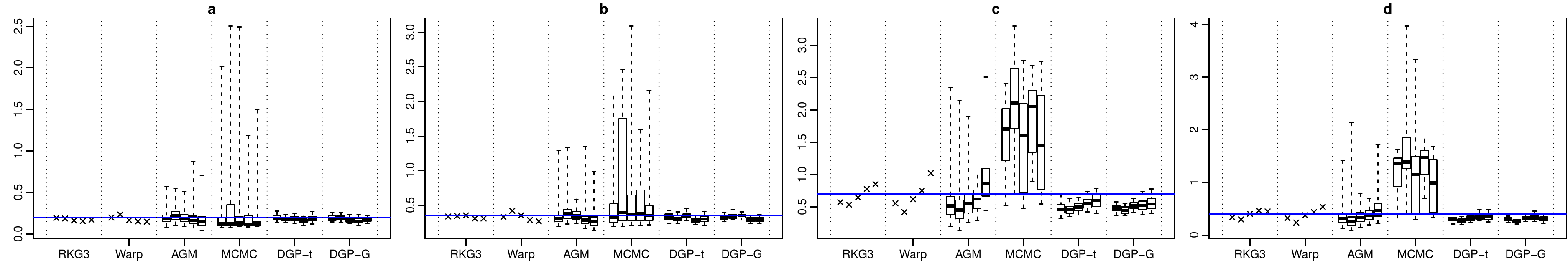}
{\bf {\scriptsize Fitz-Hugh-Nagumo }}
\includegraphics[width=\columnwidth, height=2cm]{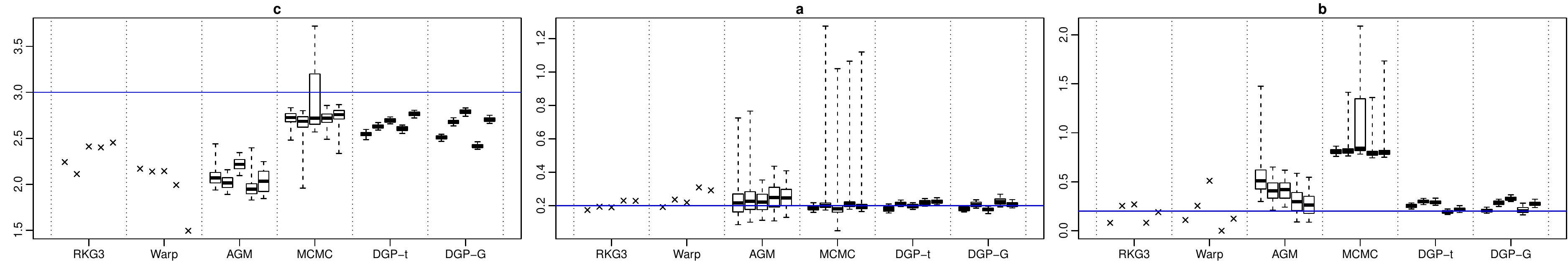} 
{\bf {\scriptsize Biopathway }}
\includegraphics[width=\columnwidth, height=2cm]{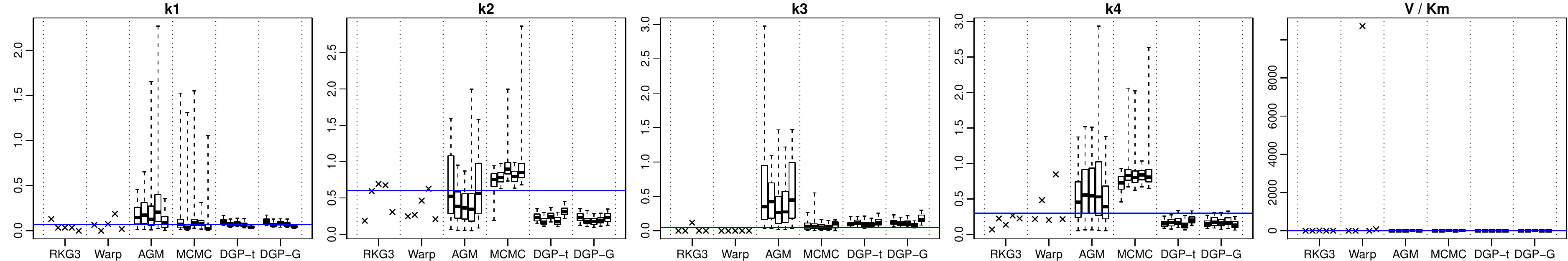}
\caption{Box-plot of posteriors over model parameters. The five box-plots for each method indicate five different repetitions of the instantiation of the noise.}
\label{fig:res0}
\end{center}
\vskip -0.2in
\end{figure*}

\begin{figure*}[h!]
\vskip 0.2in
\begin{center}
{\bf {\scriptsize Lotka-Volterra }}
\includegraphics[width=\columnwidth, height=2cm]{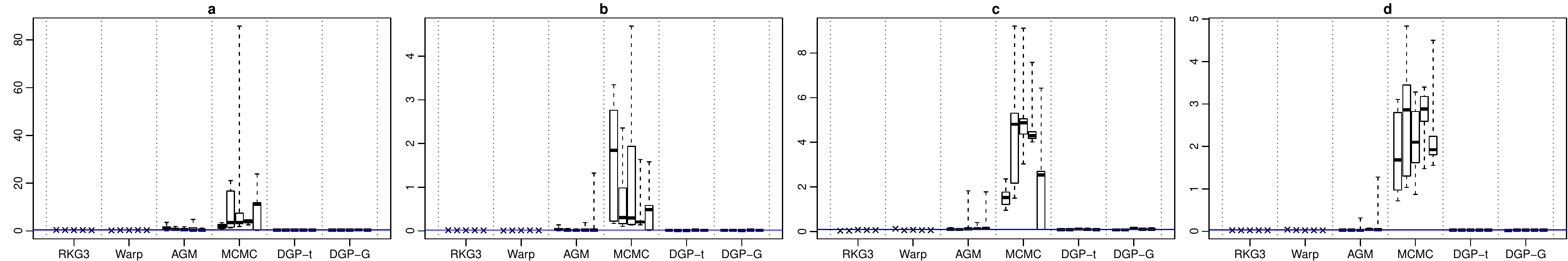}
{\bf {\scriptsize Fitz-Hugh-Nagumo }}
\includegraphics[width=\columnwidth, height=2cm]{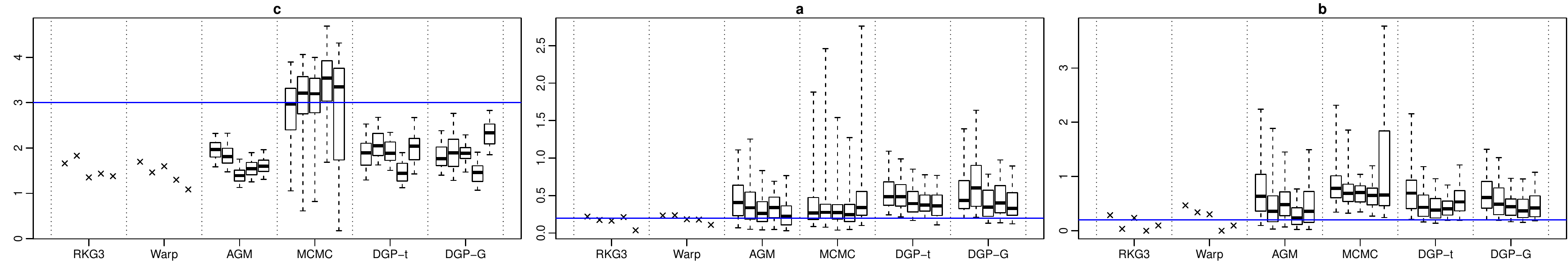} 
{\bf {\scriptsize Biopathway }}
\includegraphics[width=\columnwidth, height=2cm]{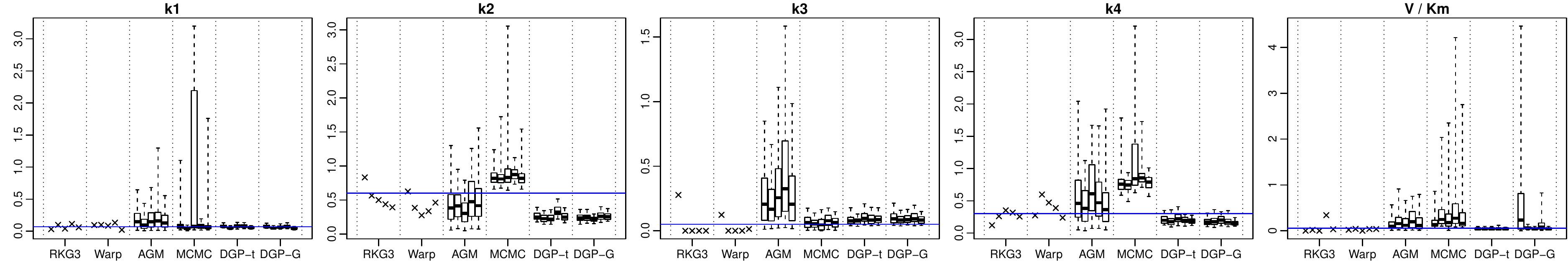}
\caption{Box-plot of posteriors over model parameters. The five box-plots for each method indicate five different repetitions of the instantiation of the noise.}
\label{fig:res1}
\end{center}
\vskip -0.2in
\end{figure*}

In figures \ref{fig:res0} and \ref{fig:res1}, we report the detailed estimate/posterior distribution obtained by the competing methods on the three \ode systems considered in this study.

\newpage 

\section{Interpolation results using Mat\'ern covariance in shallow \gps}
We report here the result of interpolating FitzHugh-Nagumo \ode with \gps with Mat\'ern covariance. 
Note that the process is still stationary, but it allows for the modeling of non-smooth functions when $\nu$ is small. 

\begin{figure*}[h!]
	\vskip 0.2in
	\begin{center}
		\includegraphics[width=\columnwidth]{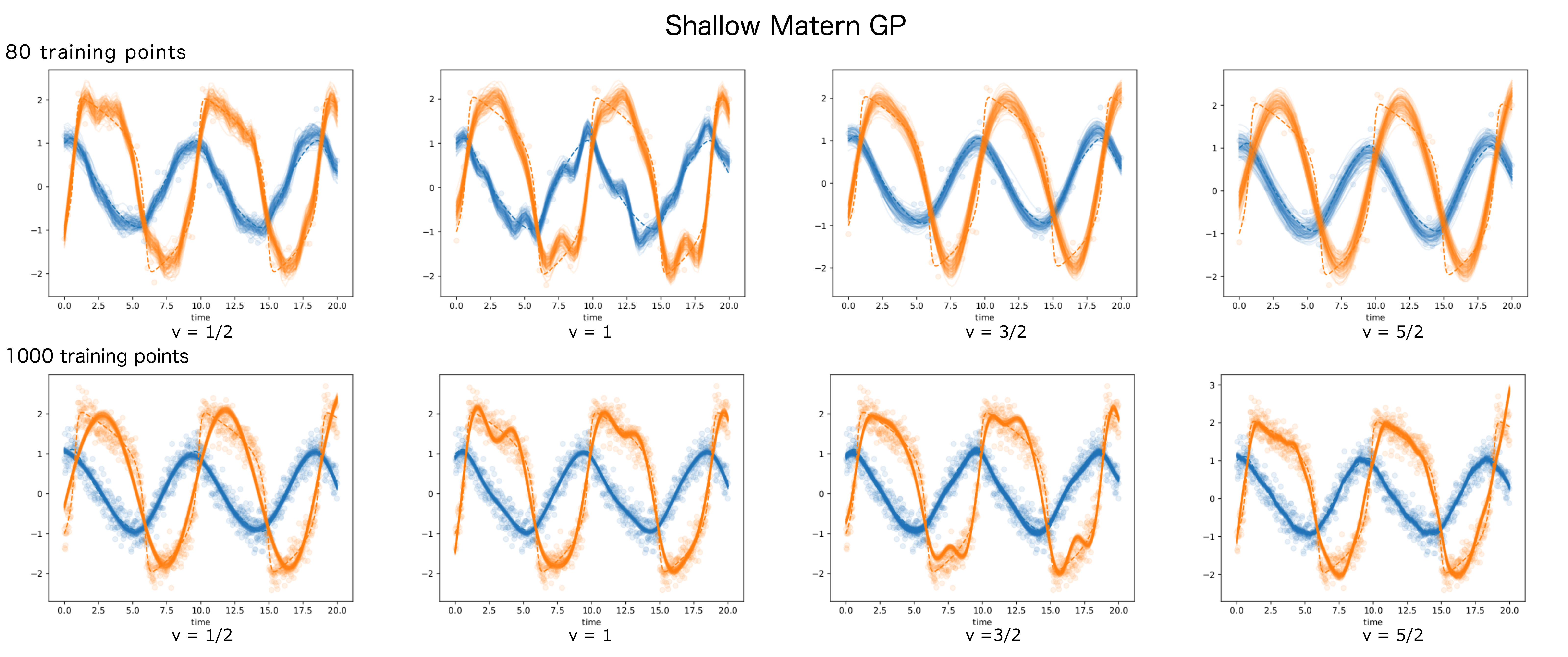}
		\caption{Modeling sampling points from the FitzHugh-Nagumo \ode with \gps with Mat\'ern covariance.}
		\label{fig:matern}
	\end{center}
	\vskip -0.2in
\end{figure*}



\end{document}